\documentclass[conference]{IEEEtran}
\IEEEoverridecommandlockouts
\usepackage{cite}
\usepackage{amsmath,amssymb,amsfonts}
\usepackage{algorithmic}
\usepackage{graphicx}
\usepackage{comment}
\usepackage{textcomp}
\usepackage{xcolor}
\usepackage{color}
\usepackage{lipsum}
\usepackage{times}
\usepackage{textcomp}
\usepackage{amsmath,amssymb,amsfonts}
\usepackage[inline]{enumitem}
\usepackage[ruled,vlined,linesnumbered]{algorithm2e}
\usepackage{color}
\usepackage{tablefootnote}
\usepackage{mathtools}

\usepackage{amsthm}
\usepackage{imakeidx}
\usepackage{float}
\usepackage{multicol}
\usepackage{booktabs} 
\usepackage{subcaption}
\usepackage{listings}
\usepackage{verbatim}

\makeatletter

\def\BibTeX{{\rm B\kern-.05em{\sc i\kern-.025em b}\kern-.08em
    T\kern-.1667em\lower.7ex\hbox{E}\kern-.125emX}}
\begin{document}

\title{FedAR: Activity and Resource-Aware Federated Learning Model for Distributed Mobile Robots}

\author{\IEEEauthorblockN{Ahmed Imteaj and M. Hadi  Amini}
\IEEEauthorblockA{\textit{School of Computing and Information Sciences, College of Engineering and Computing, Florida International University} \\
\textit{Sustainability, Optimization, and Learning for InterDependent networks laboratory (solid lab)}\\
Miami, FL, USA \\\{aimte001,  moamini\}@fiu.edu}
}
\maketitle
\begin{abstract}
Smartphones, autonomous vehicles, and the Internet-of-things (IoT) devices are considered the primary data source for a distributed network. Due to a revolutionary breakthrough in internet availability and continuous improvement of the IoT devices capabilities, it is desirable to store data locally and perform computation at the edge, as opposed to share all local information with a centralized computation agent. A recently proposed Machine Learning (ML) algorithm called \textit{Federated Learning (FL)} paves the path towards preserving data privacy, performing distributed learning, and reducing communication overhead in large-scale machine learning (ML) problems. This paper proposes an FL model by monitoring client activities and leveraging available local computing resources, particularly for resource-constrained IoT devices (e.g., mobile robots), to accelerate the learning process. We assign a trust score to each FL client, which is updated based on the client's activities. 
We consider a distributed mobile robot as an FL client with resource limitations either in memory, bandwidth, processor, or battery life. We consider such mobile robots as FL clients to understand their resource-constrained behavior in a real-world setting. We consider an FL client to be untrustworthy if the client infuses incorrect models or repeatedly gives slow responses during the FL process. After disregarding the ineffective and unreliable client, we perform local training on the selected FL clients. To further reduce the straggler issue, we enable an asynchronous FL mechanism by performing aggregation on the FL server without waiting for a long period to receive a particular client's response.



\end{abstract}


\section{Introduction}
\subsection{Motivation}
The ubiquitous nature of IoT devices causes huge data streams due to their widespread applications. Storing and processing such vast amounts of data in a centralized location is costly, highly insecure, and time-consuming. Further, in some cases, the nature of local data is sensitive and requires further security measures to ensure user confidentiality and data privacy while exchanging data for computation and decision-making purposes.
Sensitive data such as captured images, browsing history,  personal information, and location-based services can be utilized for recommendations, social advertising, prediction, or incurring any potential privacy risks. Such sensitive information is unsafe to share with the server if potential privacy issues are not checked \cite{yang2019federated}. Besides, as privacy preservation awareness is rising, legal laws and restrictions, i.e., General Data Protection Regulation (GDPR), are becoming prominent, which reduce the feasibility of data aggregation \cite{lyu2020threats}. The conventional centralized ML techniques can not be implemented in a distributed fashion due to infrastructure fallibility, including intermittent or weak network connectivity, limited bandwidth, or response delay \cite{li2018learning}. To tackle the above-mentioned issues, an alternative distributed ML paradigm named \textit{Federated Learning (FL)} is proposed in \cite{mcmahan2017communication} that performs on-device training based on client's local data, pushes client's training parameters at the edge, and learns from the global model. The popularity of FL applications is increasing due to their high-acceptability, particularly in improving user privacy. The FL process enables network clients to generate a joint ML model that enhances privacy by not exposing any clients' private data. Another important factor of FL is that it works with nonindependent and identically distributed (non-IID) data samples observed in real-world settings \cite{mcmahan2016federated, bonawitz2017practical}. That means the FL algorithm is capable of handling the changes over time in the distribution of collected data samples. However, during an FL process, the client selection is crucial as a straggler, and an unreliable client can retard the learning process and prolong the model convergence time. Any client that is selected for a training phase is called a participant. A participant may turn into a straggler if the participant is underpowered in system requirements for a particular model training. Most of the existing FL approach avoids the straggler issue by assuming all the FL clients as proficient nodes and randomly selecting clients for the training rounds. However, the presence of such straggler clients has a vital impact on the overall performance of learning model \cite{imteaj2020federated}.
Further, there is a possibility of selecting a vulnerable FL client, specifically in an FL-based IoT (FL-IoT) environment, where the devices are more prone to susceptibility. Therefore, we need to monitor the clients' available resources, behaviors, and contributions towards training a learning model. 

In our proposed approach, we consider mobile robots integrated with comparatively low processing units, limited memory, bandwidth, and battery power instead of assuming proficient clients with sufficient resource availability. We give instructions to each registered mobile robot for collecting data samples. We track the robot clients through remote sensing mechanisms, and each client gets triggered upon receiving a particular notification from the server \cite{imteaj2019distributed}. Before executing a training round, we check the resource availability following the system requirement for model training and examine each interested client's trust score, which is updated based on their previous training performance. By enabling asynchronous FL, we further reduce the straggler effect by ignoring the unresponsive clients.


\subsection{Literature Reviews}
The FL research domain is increasingly evolving due to its capability to handle non-IID data, preserve privacy, and reduce communication overhead. Prior works from various disciplines, including distributed systems, ML, databases, cryptography, and data mining, focused on improving the FL method. An overview of the FL system design is presented in \cite{bonawitz2019towards}, where they discussed the complexity of FL design by posing challenges of FL implementation and experimental evaluation. In consequence, some open-source frameworks are proposed to handle FL mechanisms such as TensorFlow Federated (TFF) designed by Google \cite{tensorflow2015-whitepaper}, Federated AI Technology Enabler (FATE) from WeBank \cite{WeBank}, LEAF \cite{caldas2018leaf}, and PySyft \cite{ryffel2018generic}.

FL strategy fits best in such applications, where we need to perform model training using sensitive data. Hence, on-device training is preferable rather than passing data to the cloud. The authors in \cite{huang2018loadaboost} proposed an FL-based model training for predicting heart attack in a patient. By configuring the necessary setup and integrating wearable devices, they observed the patient's health data and performed FL-based on-device model training. Besides, a Federated Transfer Learning (FTL) framework is presented in \cite{yang2019federated}. In that paper, they considered each hospital as an FL client and performed collaboration among neighbor hospitals to improve the FL model. In \cite{ge2020fedner}, FL-based privacy-preserving named entity recognition (NER) is implemented that can identify different medicinal attributes (e.g., drug names, its reactions, and symptoms) by analyzing medical texts and performing classification. 

Device-centric application data is utilized to build a recommendation system that aims to predict a rating for an item or user preference. For such a recommendation system, usually, one user data is shared with other users, and privacy is violated in many cases. A federated meta-learning-based recommendation system is proposed in \cite{chen2018federated} considering the privacy issues. Each FL client shares their generated algorithm with the server to carry out an effective global model. Similarly, privacy-preserving news recommendation system \cite{qi2020fedrec} and personalized recommendation system \cite{ammad2019federated} are constructed based on the FL concept. Another interesting application of FL is user-typing-behavior-based next-word prediction designed by \cite{hard2018federated}. Their application can read a user's mind while typing by analyzing the user's messages and browsing history. A similar type of application to predict emoji on a mobile keyboard is presented in \cite{ramaswamy2019federated}. Moreover, wake-word detector-based applications are prevalent nowadays that consider user voice or speech for training an FL model that can recognize the user's command. The author in \cite{leroy2019federated} designed such an FL-based wake-word detector application by using a crowd-sourced dataset that does not expose a user's voice during training. The ranking search result is another recent application of FL \cite{bonawitz2019towards, hartmann2019federated}, where the user query and preference are not shared with any other entity, and the search result is generated based on FL training. A content suggestion framework is designed on the concept of user activities, i.e., clicking or ignoring a content, which is discussed in \cite{yang2018applied}. This application tracks and stores the user-click information on content suggestions, and constructs a model accordingly. Both training and inference are performed on-device in their work, and only the model parameters are dispatched to the server. 
However, most of the existing FL-based applications assumed that the FL clients are resource-boundless and trustworthy, i.e., any FL client can be selected for the training phase. However, if we consider a real-life FL-IoT setting, the FL client is often resource-bounded and prone to vulnerability. This paper aims to consider both resource-constrained and reliability issues in an FL environment. According to the best of our knowledge, there is no existing work that examines the resource status and scrutinizing the client activity within an FL environment.

\subsection{Contributions}
The main contributions of this paper can be listed as follows:
\begin{itemize}
    \item  We propose a novel model capable of giving a reward or a punishment to each participating FL client based on their performance during a training period. 
    \item We propose a strategy to choose only capable clients before starting the FL training phase.
    \item We consider mobile robots to collect federated data in a distributed fashion and perform FL training simulation for a resource-constrained IoT environment.
    \item We consider an asynchronous FL scheme to mitigate the straggler effect on a resource-constrained FL-IoT environment.

\end{itemize}

\subsection{Organization} The rest of this paper is organized as follows. Section 2 highlights the overview of FL. Section 3 elaborately explains our novel FL framework to minimize prediction loss and obtain high accuracy. Section 4 presents our experiment results, followed by section 5, that concludes the paper.

\section{Background of Federated Learning}
Federated Learning (FL) is a distributed ML algorithm that enables on-device learning of clients within a network instead of passing raw data to the server. Based on the clients' models, or algorithms, the FL algorithm learns a single, global model. The advantage of FL method is that both federated data and models of each client remain local. In a distributed system, federated data refers to a collection of extracted data elements hosted over a cluster of devices. For instance, any applications running on a mobile device can store those application data locally, or, using distributed sensing, mobile nodes can sense sensor data and store those without sharing with a centralized location. To initialize the learning process, federated data is required, and we need to prepare a dataset which holds data from multiple sources or agents. In a typical FL scenario, we may need to deal with a very large number of user nodes, but only a small portion of them may be available at a certain time to perform training sessions. For instance, if the client nodes are mobile robots, then the devices can only be considered for a training round when they are charged, and active.

The author in \cite{mcmahan2016communication} introduced the FL term and proposed a popular FL algorithm named Federated Averaging (FedAvg). The model training of FL is done in several communication rounds until desirable model accuracy is obtained. To decrease network traffic, the training period on each client side is carried out through different batches. Further, only local model parameters or statistical summary are shared with the server, i.e., client data does not need to leave its own device.

\begin{figure}[t]
\setlength{\belowcaptionskip}{-20pt}
\begin{center}
  \includegraphics[width=0.99\linewidth]{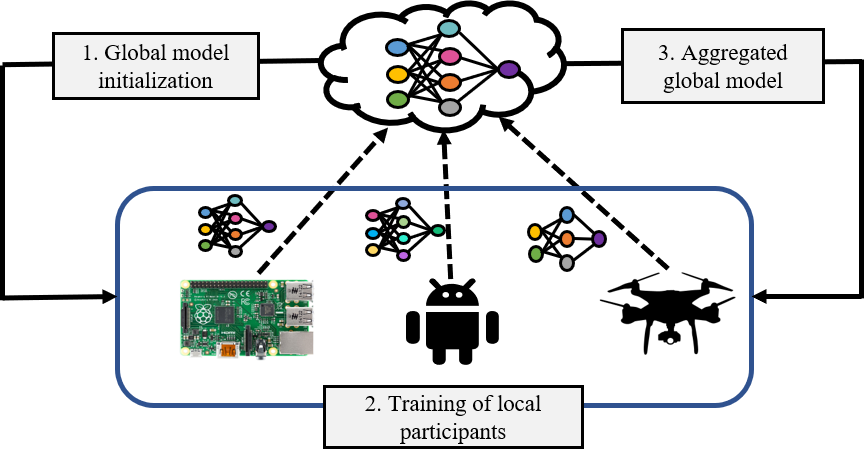}
    \caption{FL procedure considering N participants.}
    \label{fig:l}  
    \end{center}
\end{figure}

The FL method illustrated in Figure~\textbf{\ref{fig:l}} comprises 
three steps:

\noindent\emph{\textbf{Step 1 (FL Task and Global Model Initialization)}}: 
In the initial step, the FL server initializes a task with specific requirements, and an initial global model. The global model is defined by applying initial hyper-parameters and a learning rate. In each communication round, the server selects a subset of clients, known as participants. After that, the server transfers the global model to all the selected participants.

\noindent\emph{\textbf{Step 2 (On-device Training)}}: 
Each selected client performs an on-device training using their local data and updates own local model parameters by learning from the global model. The main goal of each local client is to find optimal parameters to minimize the loss function. Each client discovers the local optimal parameters by applying stochastic gradient descent (SGD) on their local available data.

\noindent\emph{\textbf{Step 3 (Perform Aggregation)}}:
The server waits to receive a local model update from each training participant. Upon receiving the model parameters, the server performs an aggregation and updates the global model. The latest global model broadcasts again to all newly selected participants. 

The local model and global model update, i.e., step 2 and step 3, are repeated until the loss function is minimized, or the model reaches a convergence.

\begin{figure}[t]
\setlength{\belowcaptionskip}{-20pt}
\begin{center}
  \includegraphics[width=0.99\linewidth]{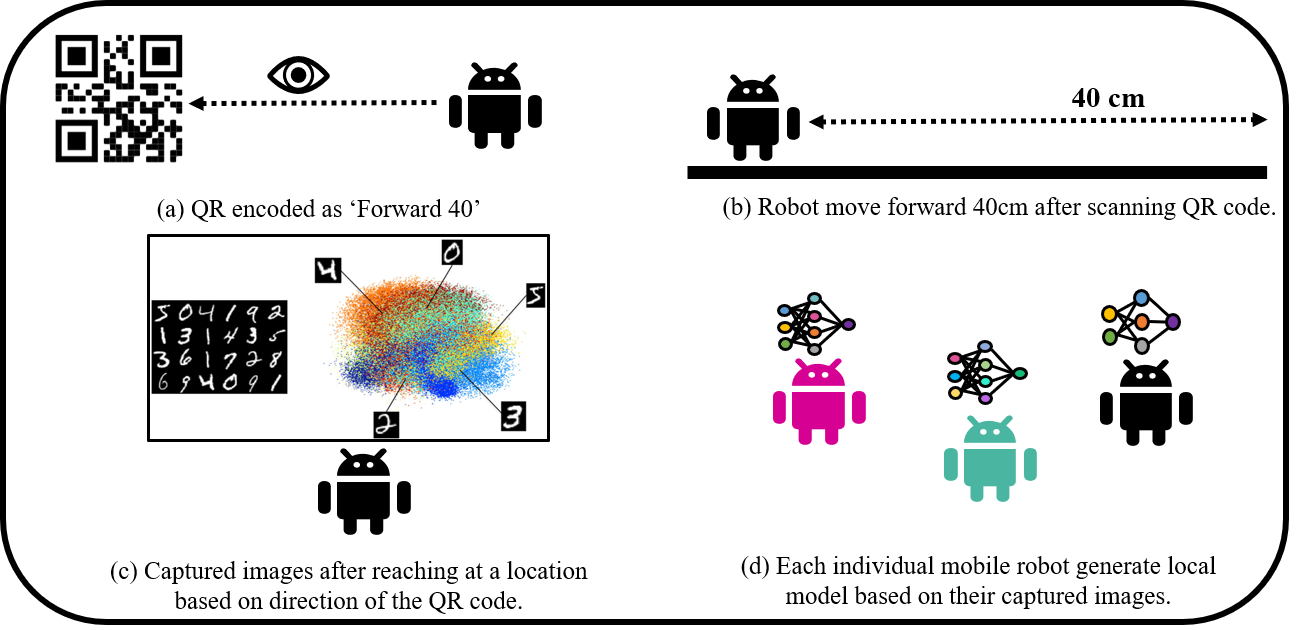}
    \caption{Real-time data collection procedure in an FL environment.}
    \label{fig:2}  
    \end{center}
\end{figure}

\section{System Description}
In this paper, we assume that we have a resource-constrained FL environment, where a client may have a straggler effect, and any interested client may provide inappropriate model information during the training process. The main reason of straggler effect is the resource shortage issue that leads us to think about a strategy to avoid an unreliable or inefficient client during the learning process.
Moreover, the existence of unreliable or inconsistent clients can prolong the convergence time and may create a negative impact on overall model accuracy. To understand the resource-constrained FL behavior, we focus on collecting real-time federated data collection, handling resource heterogeneity, and monitoring client activity during an FL process.

\subsection{Real-time Data Collection through Distributed Sensing}
In a federated dataset, we need to have a column of client id that represents which client possesses the corresponding row information. To collect federated data using distributed mobile robots (clients), we need to send a set of instructions to those clients. One naive strategy is to collect data by controlling the robots remotely. However, such a strategy is not convenient if we have many clients, and we need to keep track on each client's information. We design an approach for controlling mobile robots through which any robot can be tracked, and acknowledged about required instructions in a convenient way. In this approach, any new client interested in joining an FL network needs to commit registration by providing their credentials. We generate a token and store the generated token in our server. A token is a unique identifier that helps to recognize each robot client in a distinguishable way. The token can be used as the client id in our federated dataset. More details can be found in our prior work \cite{imteaj2019distributed}. We infuse functionalities to each robot so that they can identify a push notification when it is sent from the server. The push notification is used to trigger a robot and pass the commands as a Quick Response (QR) code. We use QR code to hide raw instructions. We enable a QR code scanning mechanism to each mobile robot to read the instructions given as a form of a QR code and understand the meaning of the QR code. After successfully scanning the QR code, a robot client can get the instructions as a form of an array. The first element of the resulting array is the key, and the following elements are the parameters. For instance, an instruction of $Forward$ $40$ $10$ $0$ means move forward with speed $40ms^{-1}$ for $10$ seconds with an angle of $0$ $degree$. In case we have multiple instructions within a QR code, we split those by a colon separator. By reaching at a desired place, the robot can capture images or can sense the environment. 
After that, the collected data are properly labeled. Each distributed robot can train themselves by the labeled data, and generate a local model that is shared with the server. In Figure \textbf{\ref{fig:2}}, we presented an IoT based data collection overview for an FL environment.

\subsection{Construction of a Trust and Resource-aware Framework}
\subsubsection{Publish FL Task}
An FL process can be considered a monopoly market, where a task publisher acts as a monopolist operator and a set of mobile robots $\mathcal{N}=\{1, \ldots, N\}$ act as the clients. Each client $n \in \mathcal{N}$ uses its local training dataset of size $s_{n}$ for being a part of the FL task. Each of the training data has an input-output pair, where the input is a vector that contains data sample features, and the output indicates a label for each input vector. The task publisher broadcasts an FL task with minimum resource requirements (e.g., memory, battery, and processing power) and a minimum trust score to qualify for the task participation. 

\subsubsection{Check Resource Availability of Interested Clients}
Upon receiving information about interested clients' resource availability, the server filters out the candidates that do not meet the task requirement. Only the robot that satisfies the requirements has a chance to join the training phase. In Algorithm \textbf{\ref{alg:1}}, we have a function $CheckResource$ that takes an input of memory ($\mathcal{M}$), bandwidth ($\mathcal{B}$), and battery life ($\mathcal{E}$) in line \textbf{14}. We generate resource availability for the three give inputs (line \textbf{15}) and compare with the task requirement $\mathcal{L}_{Req}$ (line \textbf{16}). If available resources satisfy the task requirement, then we add that client to a list $RA$ (line \textbf{17-18}).

\begin{figure*}[!htbp]
\setlength{\belowcaptionskip}{-20pt}
\begin{center}
  \includegraphics[width=0.70\linewidth]{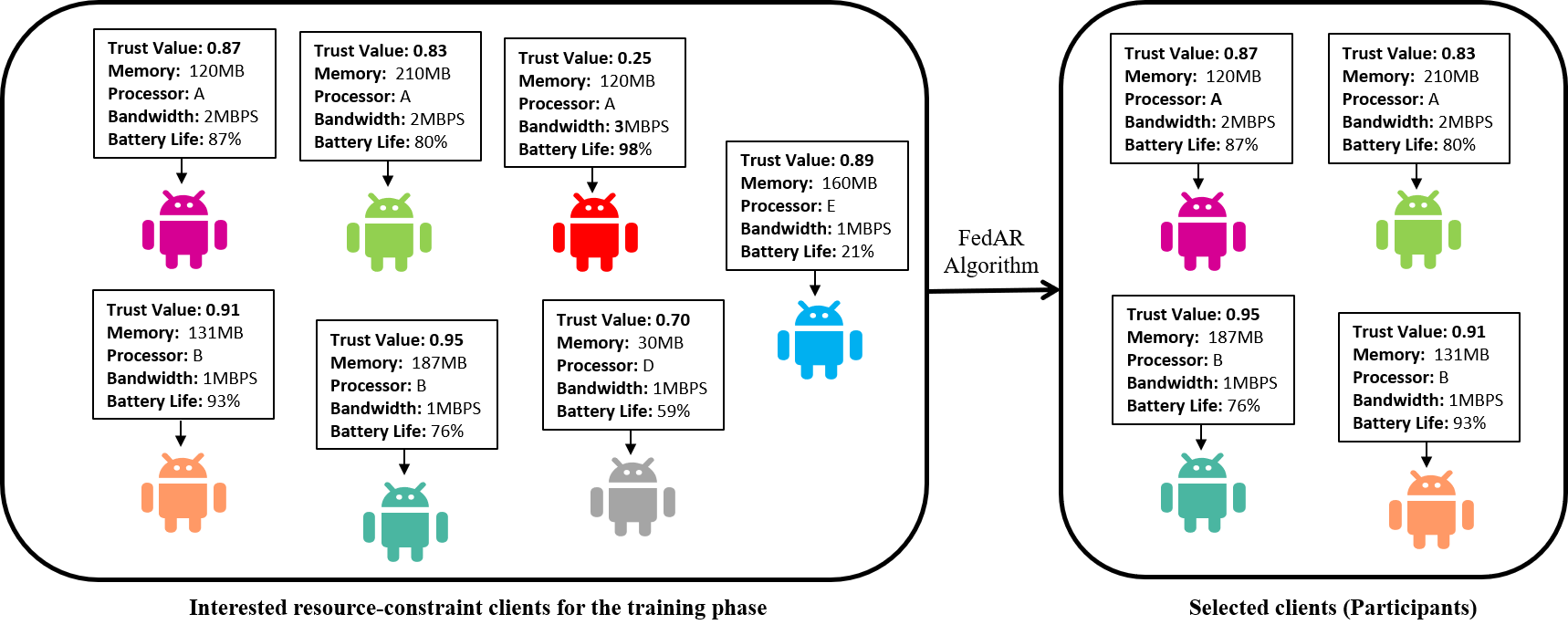}
    \caption{Selection of proficient and trustworthy clients for training phase.}
    \label{fig:3}  
    \end{center}
\end{figure*}

\subsubsection{Calculate Trust of FL-client}
In our proposed FedAR model, we choose a fraction of clients that may change at each training round according to resource availability (e.g., memory availability) and the client's updated trust value. To construct a trust model, we pass the iteration number/communication round ($i$), client id ($m$), global model parameters ($w_i$), threshold time $t$ for sending back local model, and model deviation $\gamma$. Each FL participant needs to send back their local model parameter within a given period of $t$ that is set by the task publisher. However, we may encounter various situations (e.g., fast or slow response) during the training process and may receive different FL participants' responses (e.g., inappropriate model information). If a client joins the FL network for the first time, we set its trust score as $C_{m} = 50$. We add a trust score $C_{Interested}$ with a value of $1$ to each of the clients interested in being a part of a training phase, successfully meets the trust and resource requirement for that task, but could not join the training round. We provide $C_{Interested}$ to encourage the trustworthy and capable candidate to participate in a future task. After selecting the FL client, we reward each client that accomplishes a task, and if it fails, we decrease its trust value as a punishment. 
Every participant client in a training round must submit their model within a given time because it is not feasible for the server to wait for a client for an infinite amount of time. The task publisher can set the threshold time for a task. In case an FL participant gives responses within the desired period, we provide a reward to that client ($C_{Reward}$ with a trust value of $8$). On the other hand, when an FL client fails to accomplish its task on time, we set a penalty on that client's trust score ($C_{Penalty}$ with a trust value of $-2$). We check the past performance of that client, and if the client repeatedly failed to respond on time ($20\%-50\%$ of its participation), then we add a $C_{Blame}$ trust value to that client's trust score (where $C_{Blame}$ = $-8$). In case the client's straggling effect is observed above 50\% of its overall participation, or if the client sends a model that has high deviation compare to the other client's models, we add a $C_{Ban}$ trust value to that client's trust score (where $C_{Blame}$ = $-16$). In Table \textbf{\ref{table:1}}, we presented the assigned trust score of different events. 

\begin{table}[htbp]
\begin{center}
\caption{Trust values considering different factors in an FL environment.}\label{table:1}
\begin{tabular}{|c|c|} 
\hline \textbf{Factor} & \textbf{Value} \\
\hline $C_{initial}$ & 50 \\
\hline $C_{Reward}$ & 8 \\
\hline $C_{Interested}$ & 1 \\
\hline $C_{Penalty}$ & -2 \\
\hline $C_{Blame}$ & -8 \\
\hline $C_{Ban}$ & -16 \\
\hline
\end{tabular}
\end{center}
\end{table}

The details of our trust and resource-aware model are presented in Algorithm \textbf{\ref{alg:1}}. In line \textbf{1}, we receive communication round as $i$, client id as $m$, global model parameter as $w_i$, maximum time $t$ to finish that task, and threshold of model diversity $\gamma$. The threshold time to perform a task can be changed in different iterations by the task publisher based on the client's performance. We also do not consider a fixed threshold because, in the initial period of training, a model deviation can be larger for all clients than the global model, while after some iteration, it is supposed to get reduced to some extent. After sending all the function parameters, line \textbf{2-4} checks whether a client sends back its optimized local model within a preset time $t$. If the client sends its optimized local model within time, then we set that client's unsuccessful record ($U_m^{i}$) for that training round as $0$ and give that client a reward score ($C_{Reward}$), which is added to the client's trust score. In case a client can not send back its local model within time $t$, we set that client's unsuccessful record ($U_m^{i}$) for that training round as $1$ (line \textbf{6}). We check how frequently that client encounters unsuccessful or ineffective local model generation. If that unsuccessful record indicates below 20\% successful task completion of the overall client participation history, then we add a penalty score ($C_{Penalty}$) to that client as a sign of warning (line \textbf{7-8}). In case we get more than 20\%, but less than 50\% unsuccessful record history for a participated client, then we set a blame score ($C_{Blame}$) to the existing trust score of that client (line \textbf{9-10}). Moreover, if a client is responsible for giving a slow response for more than 50\% of its overall task participation history or sending back a model with large diversity compared to other client's local models, we assign a Ban score  ($C_{Ban}$) to the existing trust score of that client (line \textbf{11-12}). Finally, in line \textbf{13}, we append the updated trust score of each client to a list.

{\fontfamily{times}\selectfont
\begin{algorithm}  
\DontPrintSemicolon
\caption{\textbf{Activity and Resource-Aware Model.} The global model of $i^{th}$ training round is represented by $G^i$, $D_m^{i}$ indicates the local model of a mobile robot $m$ on $i^{th}$ iteration, unsuccessful record of a client $m$ on iteration $i$ is denoted by ($U_m^{i}.$), trust score $C_{m}$ for $m^{th}$ client, $t$ represents timeout, and $\gamma$ indicates deviation.} \label{alg:1}
\textbf{UpdateTrustScore ($i, m, w_i, t, \gamma$):}\;
\If{$m$ sends model $w^{i}$ within $t$}
{
    set $U_{ID(m)}^{i}$ = $0$ \;
    set $C_m$ = $C_m$ + $C_{Reward}$
    
}
\Else {
    set $U_{ID(m)}^{i}$ = $1$ \;
    \If{$\frac{1}{\mathrm{i}} \sum_{\mathrm{p}=1}^{\mathrm{i}} \mathrm{U}_{\mathrm{m}}^{\mathrm{p}}$ $<$ $0.2$}
    {
    set $T_{ID(m)}^{i}$ = $C_{Penalty}$ \;
    }
    \If{$\frac{1}{\mathrm{i}} \sum_{\mathrm{p}=1}^{\mathrm{i}} \mathrm{U}_{\mathrm{m}}^{\mathrm{p}}$ $<$ $0.5$ and $\frac{1}{\mathrm{i}} \sum_{\mathrm{p}=1}^{\mathrm{i}} \mathrm{U}_{\mathrm{m}}^{\mathrm{p}}$ $\geq$ $0.2$ }
    {
    set $C_m$ = $C_m$ + $C_{Blame}$
    }
    \ElseIf{$\frac{1}{\mathrm{i}} \sum_{\mathrm{p}=1}^{\mathrm{i}} \mathrm{U}_{\mathrm{m}}^{\mathrm{p}}$ $\geq$ $0.5$ or $G^{i}$-$D_{m}^{i}$ $>$ $\gamma$
    }{
    set $C_m$ = $C_m$ + $C_{Ban}$
   }
   }
Append $C_m$ to $TrustList$

\textbf{CheckResource ($\mathcal{M}_m, \mathcal{B}_m, \mathcal{E}_m$):} \;
Resource availability, $\mathcal{R}_m$ = $f(\mathcal{M}_m, \mathcal{B}_m, \mathcal{E}_m)$ \;
Compare $\mathcal{R}_m$ with $\mathcal{L}_{Req}$ \;
\If {$\mathcal{R}_m$ satisfied $\mathcal{L}_{Req}$}
{
Add $\mathcal{R}_m$ to $RA$ list 
}
\textbf{Return} $C$ and $RA$
\end{algorithm}
}

\subsubsection{Select Client for Training Phase}
In a typical FL scenario, we may need to deal with many user nodes; however, only a small portion of them may be available at a specific time to perform training sessions. For instance, if the client nodes are mobile robots, they can be considered for training round only when charged and active. 
In our FL-IoT environment, all client data are locally available, and we need to choose a subset of clients to participate in the training process. This subset of clients is changed in each round, considering the trust score and resource availability. 
After calculating the trust value and checking out resource availability, we select the eligible candidates as participants. Each participant generates their local optimal decisions after giving consent to participate in a task according to its resource conditions and local dataset content. The effectiveness of the local model update directly depends on the local dataset accuracy \cite{shayan2018biscotti}. After selecting the clients, we can reuse them again and again until our target model reaches convergence. In Figure \textbf{\ref{fig:3}}, we presented the client selection process of our proposed FedAR method.

\subsubsection{Creating Model with Forward Pass Method}
\begin{figure}[!htbp]
\setlength{\belowcaptionskip}{-10pt}
\begin{center}
  \includegraphics[width=0.9\linewidth]{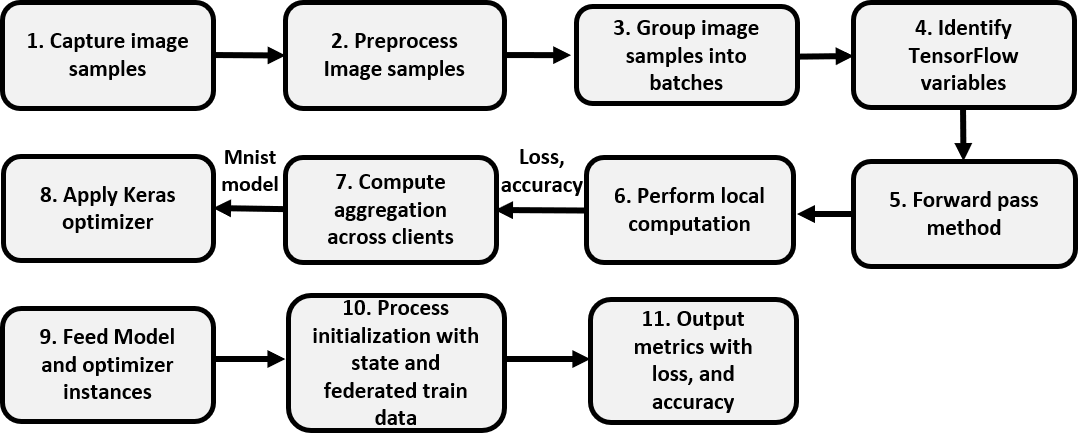}
    \caption{FL-based image classification approach considering TensorFlow framework.}
    \label{fig:4}  
    \end{center}
\end{figure}

Initially, we prepared input samples of client images to be grouped as batches. We identify the TensorFlow variables that are required in constructing our model. To represent the entire dataset, we consider a data structure that holds variables such as model weights, bias, and various counters and cumulative statistics updated during training. We define a forward pass method to compute loss with these variables and model parameters, make predictions, and update the statistics of sample batches of input. After that, we define a function that is responsible for returning a set of local metrics. Each client devices send their local metrics to the central server. The server performs aggregation upon receiving the values in a federated learning evaluation process. To the end, we build a model representation to use with TensorFlow-Federated (TFF) and apply Keras optimizer on the model. Finally, the constructed model and optimizer instances are fed to the FedAR algorithm, and the process is initialized with federated train data to generate output metrics loss and accuracy. The overall working process is depicted in Figure \textbf{\ref{fig:4}}.

\subsubsection{Evaluate Local Model Quality}
After client selection, each client can be trained with FL optimization technique (e.g., FL-SGD). The task publisher disseminates the initial shared global model parameters to all selected clients. Each mobile robot trains its model using local data and uploads its model parameters in the server. To ensure the reliability of the model update, we leverage model quality evaluation by applying FoolsGold scheme \cite{fung2018mitigating} that identifies unreliable participants by observing their local model updates' gradient diversity and helps to avoid poisoning attack. If a local client's model update performance lower than a specified threshold or a client repeatedly sent similar gradient updates, then the task publisher rejects the client update and does not update the global model. With the resource checking, unreliable client handle strategy, and malicious attacker detection approach, weak clients, and unreliable model updates are not considered during the learning process. The task publisher only considers the reliable local model update and performs a federated averaging strategy \cite{mcmahan2016communication} to generate an updated global model. 

\subsubsection{Leveraging Asynchronous FL to Accelerate Convergence}
\begin{figure}[!htb]
\setlength{\belowcaptionskip}{-13pt}
\begin{center}
  \includegraphics[height=0.65\linewidth]{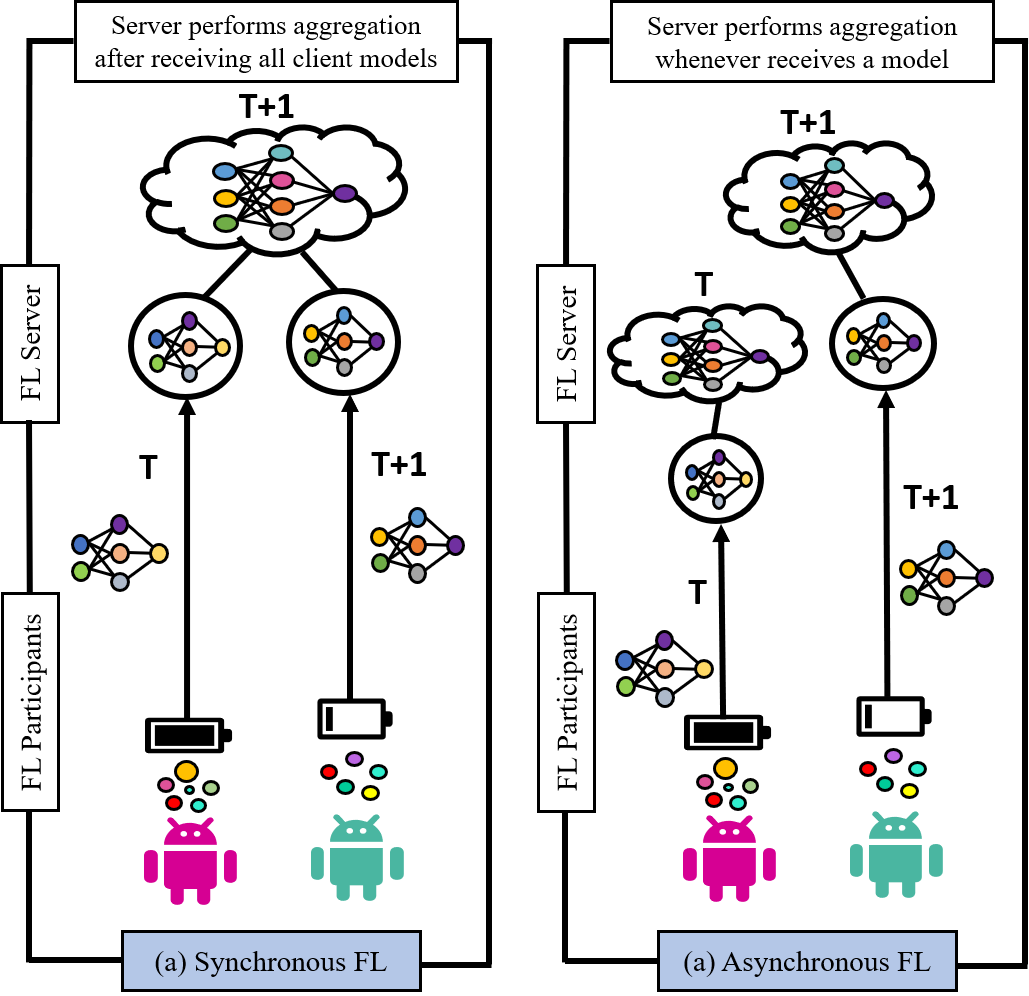}
    \caption{Synchronous and asynchronous FL approach.}
    \label{fig:asyn}  
    \end{center}
\end{figure}

This paper assumes that we have resource-constrained clients within the learning environment; therefore, performing synchronous federated learning can lead the server to wait for a long time to update the global model. In a synchronous FL, the server performs aggregation only after receiving a response from all the available clients. However, if a client has a straggler effect, then the client's slow response time can harm the overall model convergence. For such a scenario, synchronous FL does not guarantee convergence \cite{bonawitz2019towards, imteaj2020federated}. On the other hand, in asynchronous FL, the server performs aggregation every time it receives a model from a client (See Figure \textbf{\ref{fig:asyn}}), i.e., the task publisher does not have to wait for a particular client. Hence, the straggler effect does not hamper the overall model effectiveness, and it guarantees convergence \cite{lian2017asynchronous, zheng2017asynchronous, xie2019asynchronous}. We applied the asynchronous FL strategy in our proposed FedAR algorithm due to the uncertain response time of the heterogeneous resource clients. Each time the server updates its global model, the server sends the model back to all the available participants for the next iteration until the updated global model satisfies a convergence condition set by the task publisher.

\subsubsection{Update Trust Models based on Performance}
After successfully executing an FL task, the server assigns trust value to each participant based on their performances. If we update the trust value after reaching model convergence, then there remains a possibility of repeatedly selecting stragglers and unreliable clients for the training process. Rather than, we update each FL participant's trust score after completing every iteration, which eventually helps to avoid straggler or inconsistent client in the further iteration.

{\fontfamily{times}\selectfont
\begin{algorithm}  
\DontPrintSemicolon
\caption{\textbf{FedAR: Activity and Resource-aware Federated learning.} The $\mathcal{S}$ eligible clients are indexed by $u; \mathcal{B}$ is the local minibatch size, $\mathcal{F}$ is the client fraction, $E$ is the number of local epochs, $\eta$ is the learning rate, and $t$ represents timeout. }\label{alg:2}
\textbf{Registration:} Each client commits registration \;
Initialize Trust score to newly registered clients \;
Disseminate system requirements to all clients \;
\textbf{Server executes:}
initialize $w_0$  \;
Reveal resource availability by each interested client. \;
\For {each round $m=1,2, \ldots $}{ 
$RA_m$ = \textbf{CheckResource} ($\mathcal{M}_m, \mathcal{B}_m, \mathcal{E}_m$) \;

Sort available client based on $TrustList$ and $RA$, and store in $\mathcal{S}$ \;
$\mathcal{C} \leftarrow $ Top $\mathcal{S}\cdot \mathcal{F}$ clients \;
$M_{m} \leftarrow$ (random set of $\mathcal{C}$ clients)\; 
\For {each client $u \in M_{m}$ in parallel} {
$w_{m+1}^{u} \leftarrow$ ClientUpdate $\left(u, w_{m}\right)$ \; 
\If{Model received from $u$ within time $t$}{
$w_{m+1} \leftarrow w_{m+1} + \frac{n_{m}}{n} w_{m+1}^{u}$\;
}
\textbf{UpdateTrustScore ($m, u, w_m, t, \gamma$)}\;

}

}

\textbf{ClientUpdate($k, w):$} // Run on client $k$\;
$\mathcal{B} \leftarrow\left(\text {split } \mathcal{P}_{k} \text { into batches of size } \mathcal{B}\right)$ \;
\For {each local epoch $i$ from 1 to $E$} {
\For {batch $b \in \mathcal{B}$}{
$w \leftarrow w-\eta \nabla \ell(w ; b)$ }
return $w$ to server}  
\end{algorithm}
}

\subsection{Proposed FedAR Algorithm}
We presented our proposed \textit{FedAR} scheme in Algorithm \textbf{\ref{alg:2}}. Initially, each client needs to register themselves to join in a network (line \textbf{1}). We initialize each newly joined FL client with a trust value (line \textbf{2}). The server disseminates a task with a system requirement to each available client and initializes the task parameters (line \textbf{3-4}). In (line \textbf{5}), the server receives available resource information from all clients. 

In each communication round of the training phase, the available resource of each client is compared with the system requirement, and eligible clients are included in a list $ RA $ (line \textbf{6-7}). We sort the interested clients according to their trust score and resource availability, and store the eligible candidates into a list $\mathcal{S}$ (line \textbf{8}). We select a fraction of the eligible candidate in line \textbf{9}, and randomly select a subset of eligible clients (line \textbf{10}). For each selected client, we pass the latest global model (line \textbf{11-12}). We assume, there
are $M$ clients, each client have $n_u$ local data, and the overall data $n$ is partitioned among the $M$ clients with a set of indexes $\mathcal{P}_u$ on client $u$, where $n_u$ = $|\mathcal{P}_u|$. Each client splits their local data into batches, performs SGD for each of the batches, and sends back the local model parameter to the server (line \textbf{16-21}). If the server receives the client's model parameter, it immediately updates the global model by performing aggregation (line \textbf{13-14}). Finally, in line \textbf{15}, the trust score is updated based on a client's response within a threshold time of $t$.

\section{Experimented Results}
\subsection{Simulation Settings}
To create a resource-bounded real-time FL environment, we considered twelve distributed mobile robots that can follow a set of given instructions. We integrated variant sizes of memory, processor, and battery to each mobile robot to simulate the heterogeneous environment in terms of hardware configurations. We use a combination of a popular digit classification dataset called MNIST \cite{Nist} and our captured digit images using distributed mobile robots. We consider eight reliable and consistent robots and four unreliable robots. Among the four unreliable robots, two of them have issues regarding resource scarcity, while the other two generate low-quality models that can be considered a poisoning attack. We provided ten classes of digits for the unreliable workers and deliberately modified some of the training samples to mislead our FL model training process. The strength of the poisoning attack depends on the modification percentage of the labels. The robot clients use a batch size of twenty and compute five local iterations to accomplish a local SGD update. We set the same transmission rates for all the existing robot clients during the local model update to attain simplicity during execution. Therefore, we maintain the same energy consumption and transmission rates for all robot clients.

\subsection{Performance Evaluation}
We evaluate an FL setting's performance by considering different chunks of data samples for the distributed mobile robots (see Table \textbf{\ref{tab:2}}). As we mentioned before, we consider four unreliable mobile robots with resource limitations and assign fewer image samples and classes to those clients (i.e., Robot 3, Robot 5, Robot 6, Robot 9 in Table \textbf{\ref{tab:2}}). We randomly apply either Softmax or Relu activation and set instructions to each robot to collect image samples from the environment. The data label and image sample number assigned to each of the robots are presented in Table \textbf{\ref{tab:2}}. Each image is simply a matrix of pixels where each pixel indicates color density. We flatten the $28$x$28$ image samples into $784$-element arrays. After that, we shuffle the samples and group them into batches. We shuffle our images to avoid the risk of creating batches that are not representative of the overall dataset. We group the images into batches to consider mini-batch of data samples while applying local SGD. We utilize an FL framework TensorFlow 1.12.0 to evaluate the digit classification task. We apply forward pass method to perform local computation, apply a $keras$ optimizer function to obtain an optimized result, and track loss using $SparseCategoricalCrossentropy$ loss function.

\begin{figure}[!htb]
\setlength{\belowcaptionskip}{-20pt}
\begin{center}
  \includegraphics[width=0.9\linewidth]{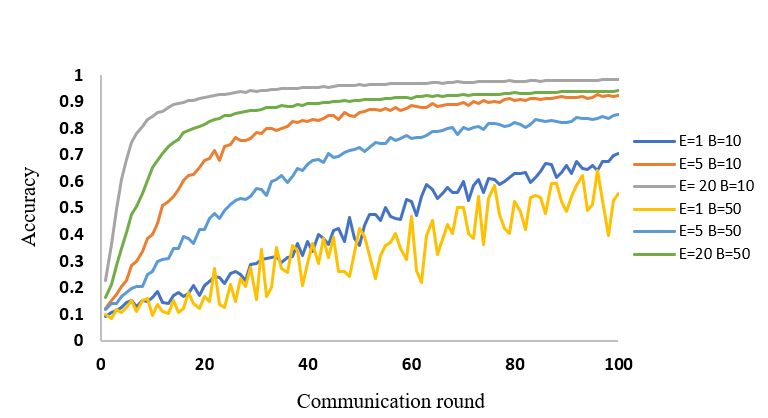}
    \caption{FL accuracy of mobile robots with variance in batch size and local epoch.}
    \label{fig:5}  
    \end{center}
\end{figure}

\begin{table}[htb!]
\begin{center}
\caption{Model architectures of FL mobile robots.}\label{tab:2}
\begin{tabular}{|c|c|c|c|} 
\hline \textbf{FL Client} & \textbf{Emnist Labels} & \textbf{Activation Function} & 
\begin{tabular}[c]{@{}l@{}}\textbf{Number of} \\ \textbf{Image Samples}\end{tabular}

\\
\hline Robot 1 & {0-9} & Softmax  & 1000 \\
\hline Robot 2 & {0-9} & ReLu &  1000 \\
\hline Robot 3  & {0,1,2,3} & Softmax & 400 \\
\hline Robot 4 & {0-9} & Softmax  & 1000 \\
\hline Robot 5  & {4,5,6} & ReLu & 300 \\
\hline Robot 6 & {7,8,9} & ReLu  & 300 \\
\hline Robot 7 & {0-9} & Softmax  & 1000 \\
\hline Robot 8 & {0-9} & ReLu  & 1000 \\
\hline Robot 9 & {5,6,8} & Softmax  & 300 \\
\hline Robot 10 & {0-9} & Softmax  & 1000 \\
\hline Robot 11 & {0-9} & ReLu  & 1000 \\
\hline Robot 12 & {0-9} & Softmax  & 1000 \\
\hline
\end{tabular}
\end{center}
\end{table}
We simulated our model's behavior with the variance of batch size and local epoch of each robot client (see Figure \textbf{\ref{fig:5}}). In Figure \textbf{\ref{fig:5}}, we use $ E $ to represent the local epoch of each client and $ B $ to indicate the data samples' batch size. We can see that the accuracy of FL increases (i.e., prediction loss decreases) as the communication round goes up. We observe that when we have a batch size of 10 and a local epoch of 20, we obtain better model accuracy for our data sample. The learning period continues until we reach a target convergence. 

\begin{figure}[!htb]
\setlength{\belowcaptionskip}{-10pt}
\begin{center}
  \includegraphics[width=0.99\linewidth]{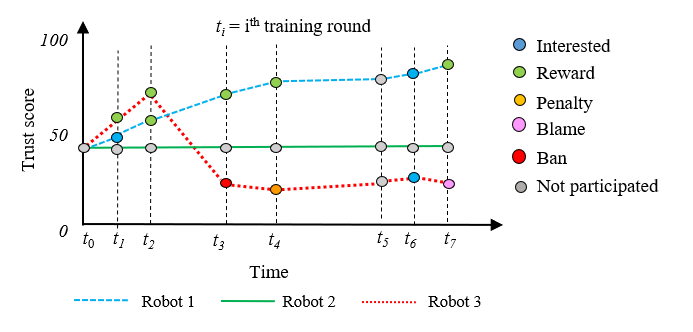}
    \caption{Activity dependent trust scores of three mobile robots.}
    \label{fig:6}  
    \end{center}
\end{figure}

\begin{figure}[!htb]
\setlength{\belowcaptionskip}{-20pt}
\begin{center}
  \includegraphics[width=0.9\linewidth]{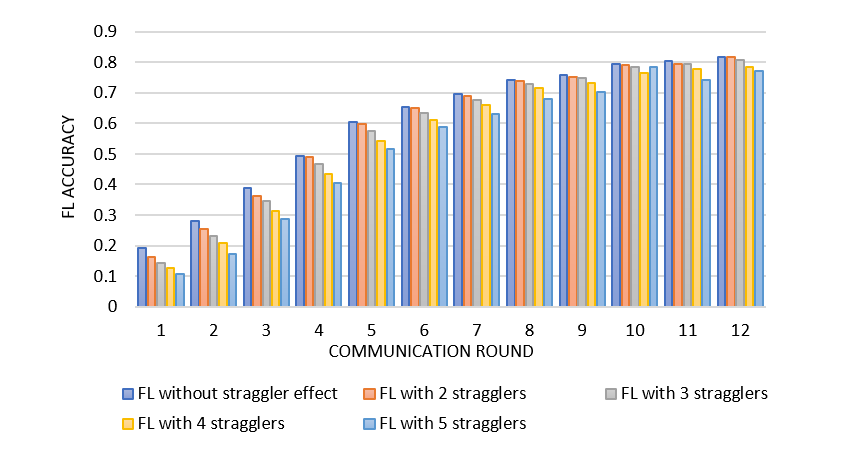}
    \caption{FL performance in presence of straggler effect.}
    \label{fig:7}  
    \end{center}
\end{figure}

To observe the clients' trust score update, we considered different events, e.g., successful task completion, improper model parameter infusion, delay in sending a response, interested to be a part of a training phase, and not able to participate due to unavailable resources. In Figure \textbf{\ref{fig:6}}, we visualized the trust score update of three different mobile robots in different periods.
Besides, we simulated the straggler effect on the FL process by considering different straggler robots that can not send back their local model update within a given time and eventually reduce the global model's overall accuracy. Figure \textbf{\ref{fig:7}} depicted that less number of straggler robots accelerates the FL accuracy.

\section{Conclusion}
This paper proposes a trust and resource-aware FL framework to deal with the untrustworthy and resource-constrained FL environment. For our simulation settings, we consider distributed mobile robots that have reliability and resource limitation issues. Instead of assuming all the FL clients are consistent and resource-efficient, we consider additional steps to check each client's resources and previous training performance. We enable a feature of assigning a trust score to each robot client, and avoiding the inconsistent and inadequate resource clients from the training process. To further handle the straggler effect, we selected the most proficient and reliable clients for an FL task and applied asynchronous FL to reduce the convergence time.

\bibliographystyle{IEEEtran}
\bibliography{ref}

\end{document}